\documentclass{article}

\PassOptionsToPackage{numbers, compress}{natbib}

\usepackage[final]{neurips_2020}

\usepackage[utf8]{inputenc} 
\usepackage[T1]{fontenc}    
\usepackage{hyperref}       
\usepackage{url}            
\usepackage{booktabs}       
\usepackage{amsfonts}       
\usepackage{nicefrac}       
\usepackage{microtype}      
\usepackage{amsmath}
\usepackage{appendix}
\usepackage{graphicx}
\title{Resolution Dependent GAN Interpolation for Controllable Image Synthesis Between Domains}

\author{
  Justin N. M. Pinkney \\
  \texttt{https://www.justinpinkney.com} \\
  \texttt{justinpinkney@gmail.com} \\
  \And
  Doron Adler \\
  \texttt{doronadler@gmail.com} \\
}

\begin{document}

\maketitle

\begin{abstract}
  GANs can generate photo-realistic images from the domain of their training data. However, those wanting to use them for creative purposes often want to generate imagery from a truly novel domain, a task which GANs are inherently unable to do. It is also desirable to have a level of control so that there is a degree of artistic direction rather than purely curation of random results. Here we present a method for interpolating between generative models of the StyleGAN architecture in a resolution dependent manner. This allows us to generate images from an entirely novel domain and do this with a degree of control over the nature of the output.
\end{abstract}

\begin{figure}[!h]
  \centering
  \includegraphics[width=\linewidth]{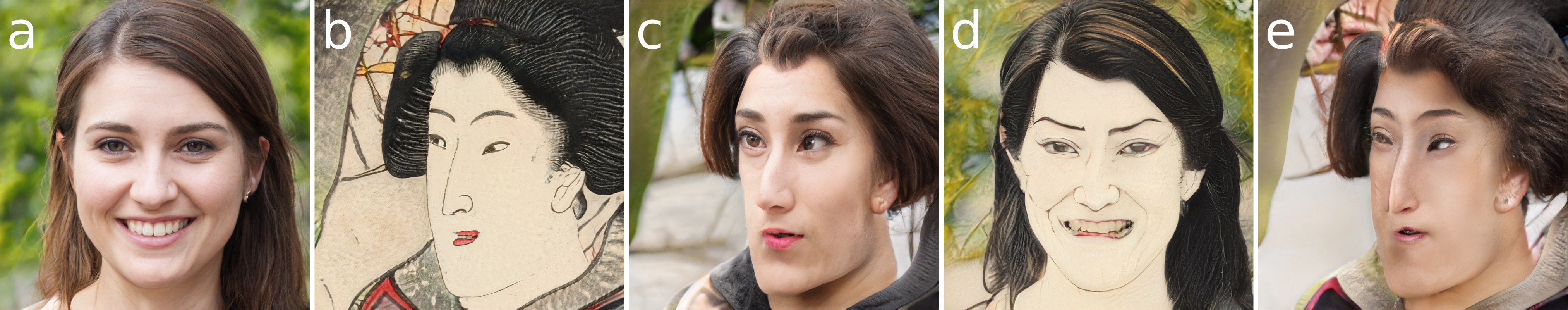}
  \caption{Interpolation between a base model trained on FFHQ (a) and a transferred model trained on ukiyo-e faces (b). Resolution dependent model interpolation creates new models (c-e) which generate images from a novel domain (we use the simple layer swapping formulation described in Section \ref{method}). Depending on which model the high and low resolution layers are taken from, we can select either structural characteristics of ukiyo-e faces with photo-realistic rendering (c) or vice-versa (d). Furthermore, by selecting which layers are swapped we can tune the effect (e).}
  \label{fig:ukiyoe}
\end{figure}

\section{Introduction}

The training stability of the StyleGAN architecture along with the availability of high quality pre-trained models \cite{pinkney_awesome_stylegan}, has made it possible for creatives and artists to produce high-quality generative models with access to only limited computing resources by using transfer learning. A model which has been generated by transfer learning, and the original "base model" have a close relationship \cite{fort2020deep} and it has been shown that linearly interpolating between the weights of two such generative models produces outputs which are approximately an interpolation of the two learned domains \cite{wang2018esrgan}. \footnote{Much experimentation on StyleGAN interpolation has been shared informally by the community on Twitter but there are no formal academic publications of these methods as far as the authors are aware.}

Simply applying linear interpolation between all the parameters in the model does not make use of a highly important element of control in StyleGAN, namely that different resolution layers in the model are responsible for different features in the generated image \cite{karras2019stylebased} (e.g. low resolutions control head pose, high resolutions control lighting). By interpolating between parameters from different models, but doing this dependent on the resolution of the particular layer, we can select and blend features from the different generators as we desire. For example, taking the pose and head shape of ukiyo-e style portraits but preserving the photo-realistic rendering, or vice-versa, as shown in Figure \ref{fig:ukiyoe}.

\section{Method}
\label{method}

\begin{figure}
    \centering
    \includegraphics[width=0.8\linewidth]{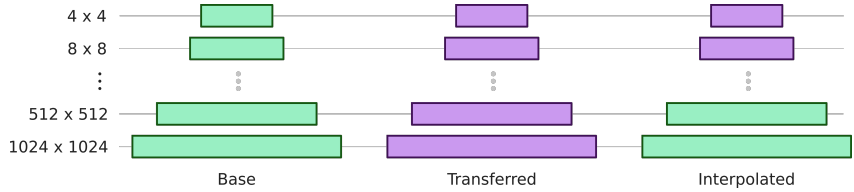}
    \caption{Schematic of the "layer swapping" interpolation scheme described in Section \ref{method}. Each block represents a resolution level in StyleGAN, the final interpolated model is composed of blocks taken from each of the input models depending on the resolution (not all blocks are shown for brevity).}
    \label{fig:diagram}
\end{figure}

\begin{enumerate}
    \item Start with a pre-trained model with weights $p_{base}$, the Base model.
    \item Train the model on a new dataset to create a model via transfer learning with weights $p_{transfer}$, the Transferred model.
    \item Combine the weights from both original and new Generators into a new set of weights $p_{interp}$.
    The function used to combine the weights between two models should be dependent on the resolution $r$ of the convolutional layers which are being combined. The choice of function used is arbitrary but here we use a simple binary choice between weights of each model which we term "layer swapping" (for more details see Appendix \ref{mathdetails}).
    \begin{align}
        p_{interp} &= (1-\alpha) p_{base} + \alpha p_{transfer} \\
        \alpha &= \begin{cases} 1, \text{where } r<=r_{swap} \\ 0, \text{where } r>r_{swap}\end{cases}
    \end{align}
    where $r_{swap}$ is the resolution level at which the transition from one model to another occurs.
    \item The new weights $p_{interp}$ are then used to create the Interpolated model.
\end{enumerate}

\section{Results - Toonification}

\begin{figure}
  \centering
  \includegraphics[width=\linewidth]{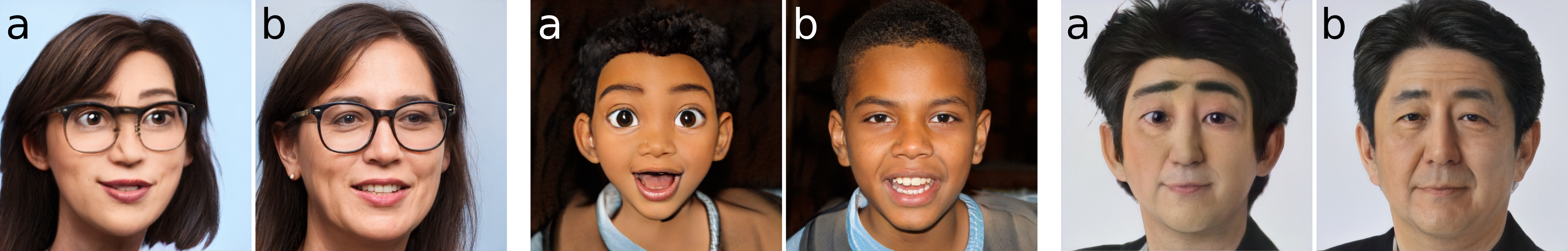}
  \caption{The interpolated model (a) produces images with the structural characteristics of a cartoon, but with photo-realistic rendering. When comparing the same latent vector input to the original FFHQ  model (b) the identity appears largely preserved thus the interpolated model gives a "cartoonification" effect. The right most pair shows a result after encoding an image of Shinzo Abe.}
  \label{fig:toonify}
\end{figure}

We further demonstrate resolution dependent interpolation for the generation of photo-realistic faces exhibiting the structural characteristics of a cartoon character. We combine high resolution layers of an FFHQ model with the low resolution layers from a model transferred to animated character faces. This gives the appearance of realistic facial textures with the structural characteristics of a cartoon (e.g. large eyes, small chin). When given the same latent vector as input, the base and interpolated models generate faces with many of the same broad characteristics in terms of identity. We can then use the well established practice of encoding an arbitrary face into the base model \cite{karras2020analyzing}\cite{puzer_encoder} and use this latent vector as input to the interpolated model to give a "Toonified" version of the original image, see Figure \ref{fig:toonify}.

\begin{ack}
No external funding was received for this work.
\end{ack}

\small

\bibliography{lib}

\begin{thebibliography}{13}
\providecommand{\natexlab}[1]{#1}
\providecommand{\url}[1]{\texttt{#1}}
\expandafter\ifx\csname urlstyle\endcsname\relax
  \providecommand{\doi}[1]{doi: #1}\else
  \providecommand{\doi}{doi: \begingroup \urlstyle{rm}\Url}\fi

\bibitem[Abdal et~al.(2019)Abdal, Qin, and Wonka]{abdal2019image2stylegan}
R.~Abdal, Y.~Qin, and P.~Wonka.
\newblock Image2stylegan: How to embed images into the stylegan latent space?,
  2019.

\bibitem[Aizawa et~al.(2020)Aizawa, Fujimoto, Otsubo, Ogawa, Matsui, Tsubota,
  and Ikuta]{Aizawa_2020}
K.~Aizawa, A.~Fujimoto, A.~Otsubo, T.~Ogawa, Y.~Matsui, K.~Tsubota, and
  H.~Ikuta.
\newblock Building a manga dataset “manga109” with annotations for
  multimedia applications.
\newblock \emph{IEEE MultiMedia}, 27\penalty0 (2):\penalty0 8–18, Apr 2020.
\newblock ISSN 1941-0166.
\newblock \doi{10.1109/mmul.2020.2987895}.
\newblock URL \url{http://dx.doi.org/10.1109/MMUL.2020.2987895}.

\bibitem[Collins et~al.(2020)Collins, Bala, Price, and
  Süsstrunk]{collins2020editing}
E.~Collins, R.~Bala, B.~Price, and S.~Süsstrunk.
\newblock Editing in style: Uncovering the local semantics of gans, 2020.

\bibitem[Fort et~al.(2020)Fort, Hu, and Lakshminarayanan]{fort2020deep}
S.~Fort, H.~Hu, and B.~Lakshminarayanan.
\newblock Deep ensembles: A loss landscape perspective, 2020.

\bibitem[Gabbay and Hoshen(2019)]{gabbay2019style}
A.~Gabbay and Y.~Hoshen.
\newblock Style generator inversion for image enhancement and animation, 2019.

\bibitem[Karras et~al.(2019)Karras, Laine, and Aila]{karras2019stylebased}
T.~Karras, S.~Laine, and T.~Aila.
\newblock A style-based generator architecture for generative adversarial
  networks, 2019.

\bibitem[Karras et~al.(2020)Karras, Laine, Aittala, Hellsten, Lehtinen, and
  Aila]{karras2020analyzing}
T.~Karras, S.~Laine, M.~Aittala, J.~Hellsten, J.~Lehtinen, and T.~Aila.
\newblock Analyzing and improving the image quality of stylegan, 2020.

\bibitem[Nikitko(2019)]{puzer_encoder}
D.~Nikitko.
\newblock Stylegan – encoder for official tensorflow implementation.
\newblock \url{https://github.com/Puzer/stylegan-encoder/}, 2019.

\bibitem[Pinkney(2020{\natexlab{a}})]{pinkney2020ukiyoe}
J.~N.~M. Pinkney.
\newblock Aligned ukiyo-e faces dataset.
\newblock \url{https://www.justinpinkney.com/ukiyoe-dataset},
  2020{\natexlab{a}}.

\bibitem[Pinkney(2020{\natexlab{b}})]{pinkney_awesome_stylegan}
J.~N.~M. Pinkney.
\newblock Awesome pretrained stylegan2.
\newblock \url{https://github.com/justinpinkney/awesome-pretrained-stylegan2/},
  2020{\natexlab{b}}.

\bibitem[Richardson et~al.(2020)Richardson, Alaluf, Patashnik, Nitzan, Azar,
  Shapiro, and Cohen-Or]{richardson2020encoding}
E.~Richardson, Y.~Alaluf, O.~Patashnik, Y.~Nitzan, Y.~Azar, S.~Shapiro, and
  D.~Cohen-Or.
\newblock Encoding in style: a stylegan encoder for image-to-image translation,
  2020.

\bibitem[Wang et~al.(2018)Wang, Yu, Wu, Gu, Liu, Dong, Loy, Qiao, and
  Tang]{wang2018esrgan}
X.~Wang, K.~Yu, S.~Wu, J.~Gu, Y.~Liu, C.~Dong, C.~C. Loy, Y.~Qiao, and X.~Tang.
\newblock Esrgan: Enhanced super-resolution generative adversarial networks,
  2018.

\bibitem[Zhu et~al.(2020)Zhu, Shen, Zhao, and Zhou]{zhu2020indomain}
J.~Zhu, Y.~Shen, D.~Zhao, and B.~Zhou.
\newblock In-domain gan inversion for real image editing, 2020.

\end{thebibliography}

\appendix
\appendixpage
\section{Mathematical formulation} \label{mathdetails}

Formally the method for performing this resolution dependent GAN interpolation is described in this section. If we have a particular neural network architecture which acts as our Generator function $G$ then an image $I$ can be generated by applying that function to a latent vector $z$

$$I = G(z, p_{base})$$

$p_{base}$ are the learned parameters of the model and dictate what domain of images are generated. We take a pre-trained model with parameters $p_{base}$ as an initialisation then train on a new dataset (i.e. transfer learning). This gives us a new set of parameters $p_{transfer}$ which generate images from the new domain:

$$I' = G(z, p_{transfer})$$

We then combine the parameters from the base and transferred model to give a new set $p_{interp}$ which will generate images from a domain qualitatively in-between the base and transfer datasets.

The function for combining the sets of parameters is dependent on the resolution block from which those parameters are selected. It could be an arbitrary function but here we use the simple case where it is 0 or 1 depending on the resolution, $r$, of the hidden layer activations at a given convolutional layer.

\begin{align}
    p_{interp} &= f(p_{base}, p_{transfer}, r) \\
    p_{interp} &= (1-\alpha) p_{base} + \alpha p_{transfer} \\
    \alpha &= \begin{cases} 1, \text{where } r<=r_{swap} \\ 0, \text{where } r>r_{swap}\end{cases}
\end{align}

where $r_{swap}$ is the resolution level at which the transition from one model to another occurs. The interpolated parameters $p_{interp}$ can then be used in a new model to generate images from a novel domain:

$$I'' = G(z, p_{interp}).$$

\section{Technical details}

\subsection{Experimental details}

\subsubsection{Ukiyo-e}

The dataset for training the Ukiyo-e portrait model was collected from online museum images of Japanese ukiyo-e prints. Faces were extracted and landmarks detected using Amazon Rekognition. The face alignment procedure used in for the FFHQ dataset was applied to the ukiyo-e portraits \cite{karras2019stylebased}. Images below the final resolution of 1024x1024 were upscaled using ESRGAN \cite{wang2018esrgan} with a model trained on the Manga109 dataset \cite{Aizawa_2020}. The final dataset consisted of approximately 5000 images and is publically available for reuse at \url{https://www.justinpinkney.com/ukiyoe-dataset/} \cite{pinkney2020ukiyoe}.

For transfer learning the model was initialised with weights from the config-e 1024x1024 FFHQ model. This was trained on the new dataset at a learn rate of 0.002 with mirror augmentation. Training was conducted for 312 thousand images with the default settings for StyleGAN2 FFHQ, after which the exponentially weighted average model was used for interpolation.

For model interpolation using layer swapping we set $r_{swap}=16$ for the results shown in Figure \ref{fig:ukiyoe} c and d, and $r_{swap}=32$ in panel e. For panels c and e the base model provides high-resolution layers and the transferred model provides low-resolution layers, this is reversed in panel e.

\subsubsection{Toonification}

The dataset for training the Toonification model was obtained from online images and the faces detected and aligned using dlib and the FFHQ procedure for alignment. The final dataset consisted of approximately 300 images.

For transfer learning the model was initialised with weights from the config-f 1024x1024 FFHQ model. This was trained on the new dataset at a learn rate of 0.002 with mirror augmentation. Training was conducted for 32 thousand images with the default settings for StyleGAN2 FFHQ, after which the exponentially weighted average model was used for interpolation.

\subsection{Model interpolation}

When performing model interpolation we divide the StyleGAN architecture into sets of layers according to the spatial resolution of the activation passing through the network at that point. These range from 4x4 to 1024x1024 in the FFHQ model presented in \citet{karras2020analyzing}. The StyleGAN architecture also contains learnable parameters which are not resolution dependent in the mapping network. In this work we choose to leave the parameters of the mapping network to be equal to that of the base model. We find that using the parameters of the transferred network for these layers makes very little difference as in practice the difference in weights between the two networks for the mapping layers is very small due to the reduced learning rate of these layers.

\subsection{Arbitrary face encoding}

To find the latent vector corresponding to an arbitrary face we use an adapted version of the original projector introduced in \citet{karras2020analyzing} provided by Robert Luxemburg (https://github.com/rolux/stylegan2encoder). There are however many methods for encoding an image into the latent space of StyleGAN \cite{abdal2019image2stylegan} \cite{zhu2020indomain} \cite{gabbay2019style} \cite{richardson2020encoding} which could also be used.

\subsection{Uncurated results}

Uncurated results for the four interpolated models presented in this paper are shown below.

\begin{figure}[!h]
  \centering
  \includegraphics[width=\linewidth]{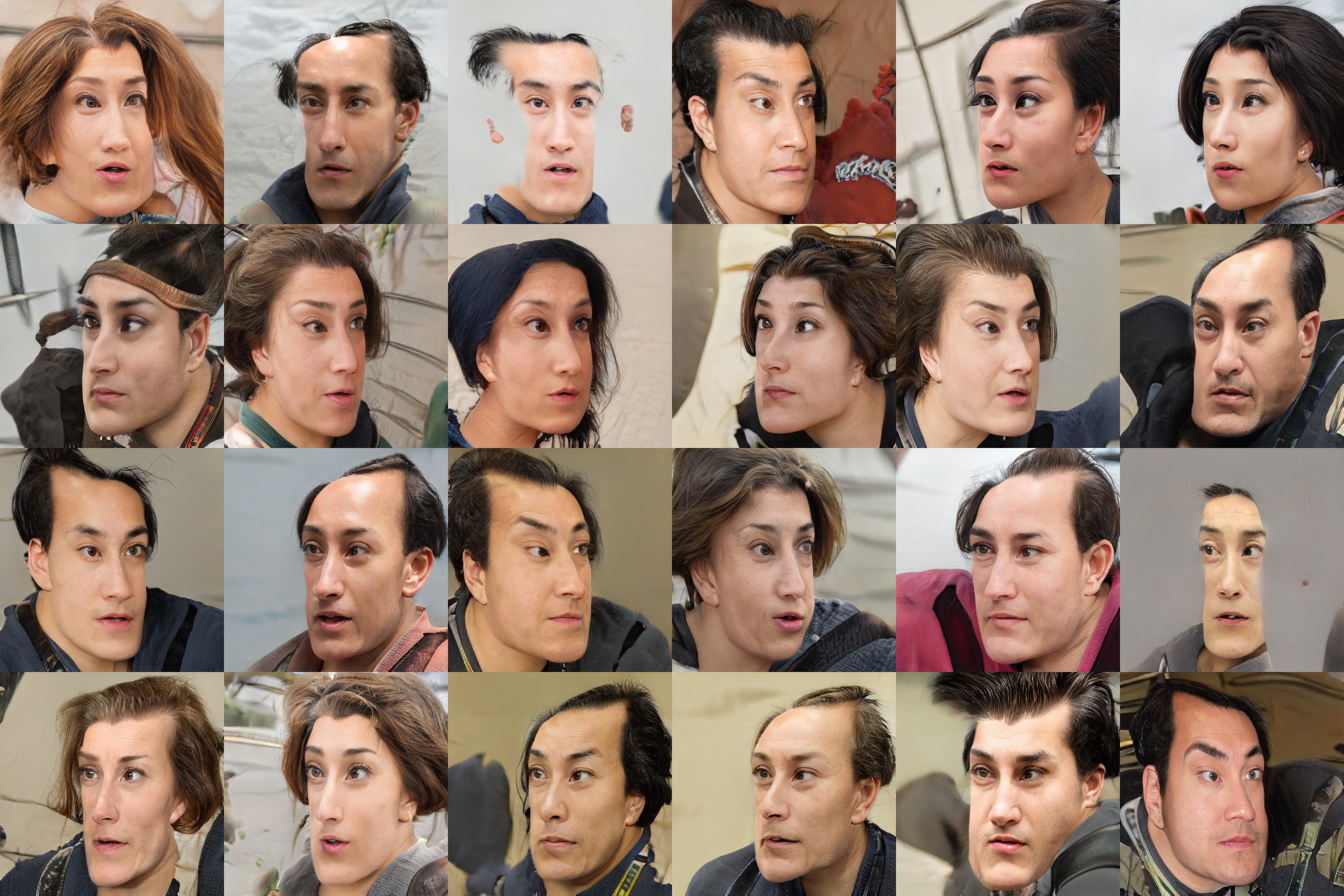}
  \caption{24 uncurated examples from the blended model shown in Figure \ref{fig:ukiyoe}c}
\end{figure}

\begin{figure}[!h]
  \centering
  \includegraphics[width=\linewidth]{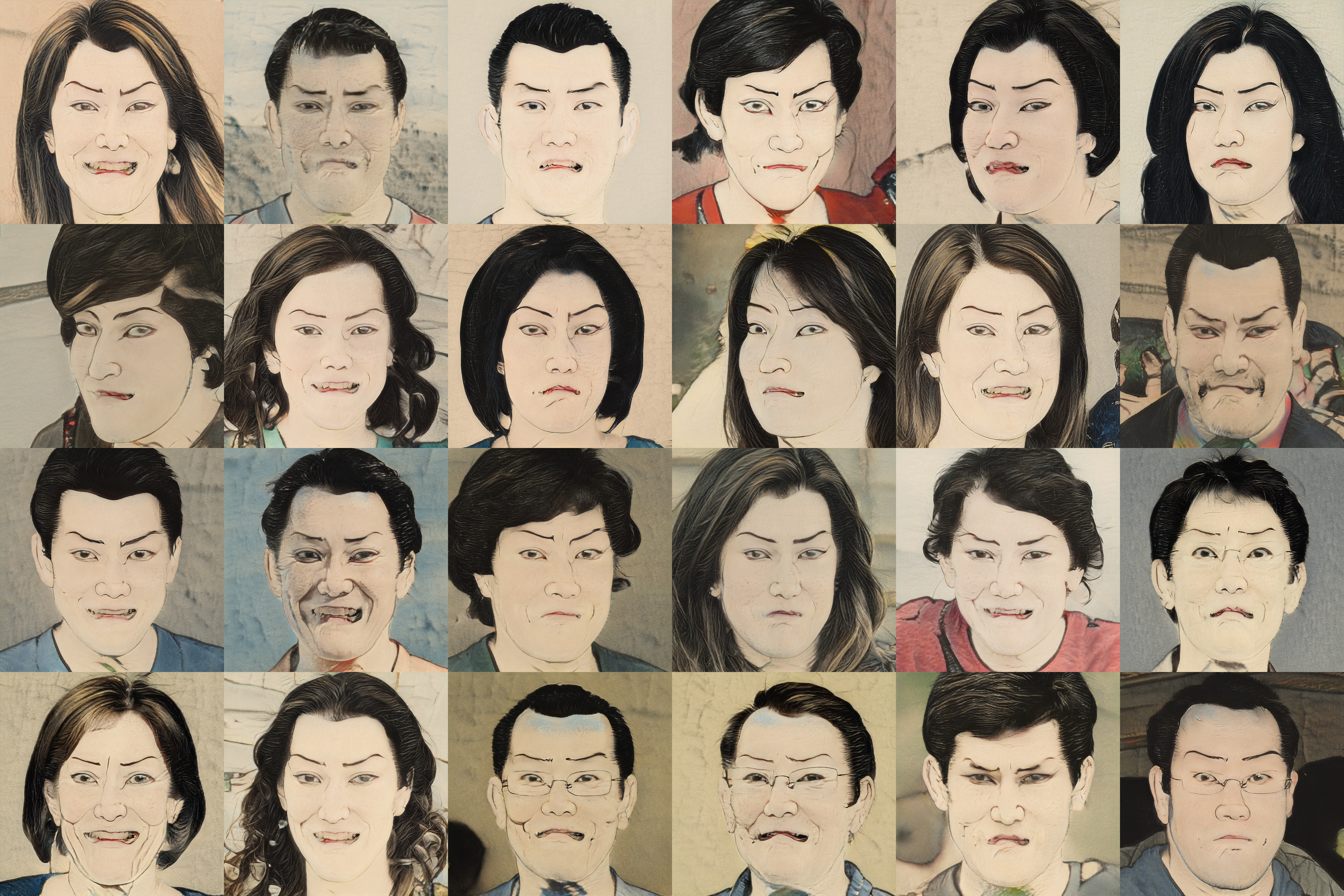}
  \caption{24 uncurated examples from the blended model shown in Figure \ref{fig:ukiyoe}d}
\end{figure}

\begin{figure}[!h]
  \centering
  \includegraphics[width=\linewidth]{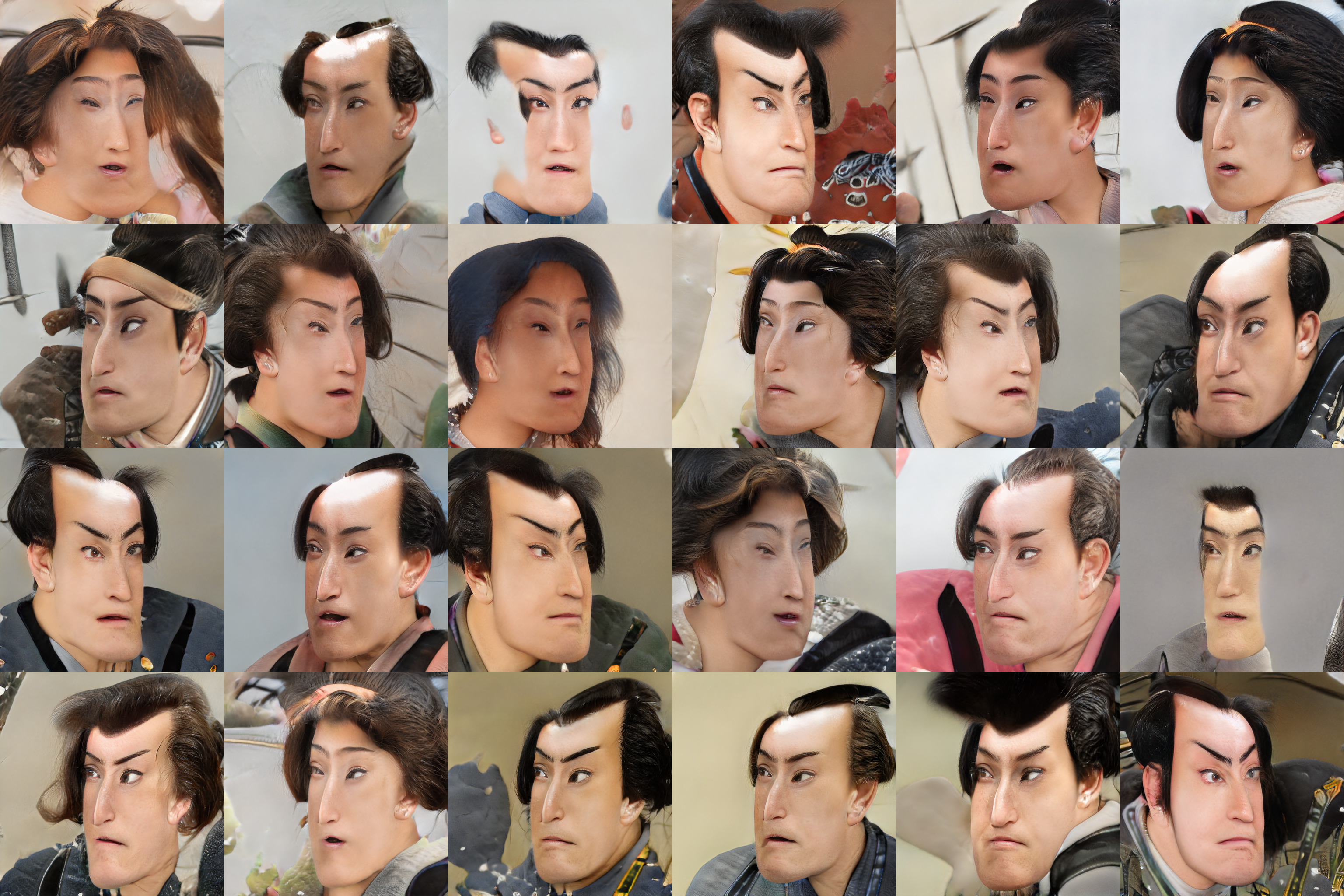}
  \caption{24 uncurated examples from the blended model shown in Figure \ref{fig:ukiyoe}e}
\end{figure}

\begin{figure}[!h]
  \centering
  \includegraphics[width=\linewidth]{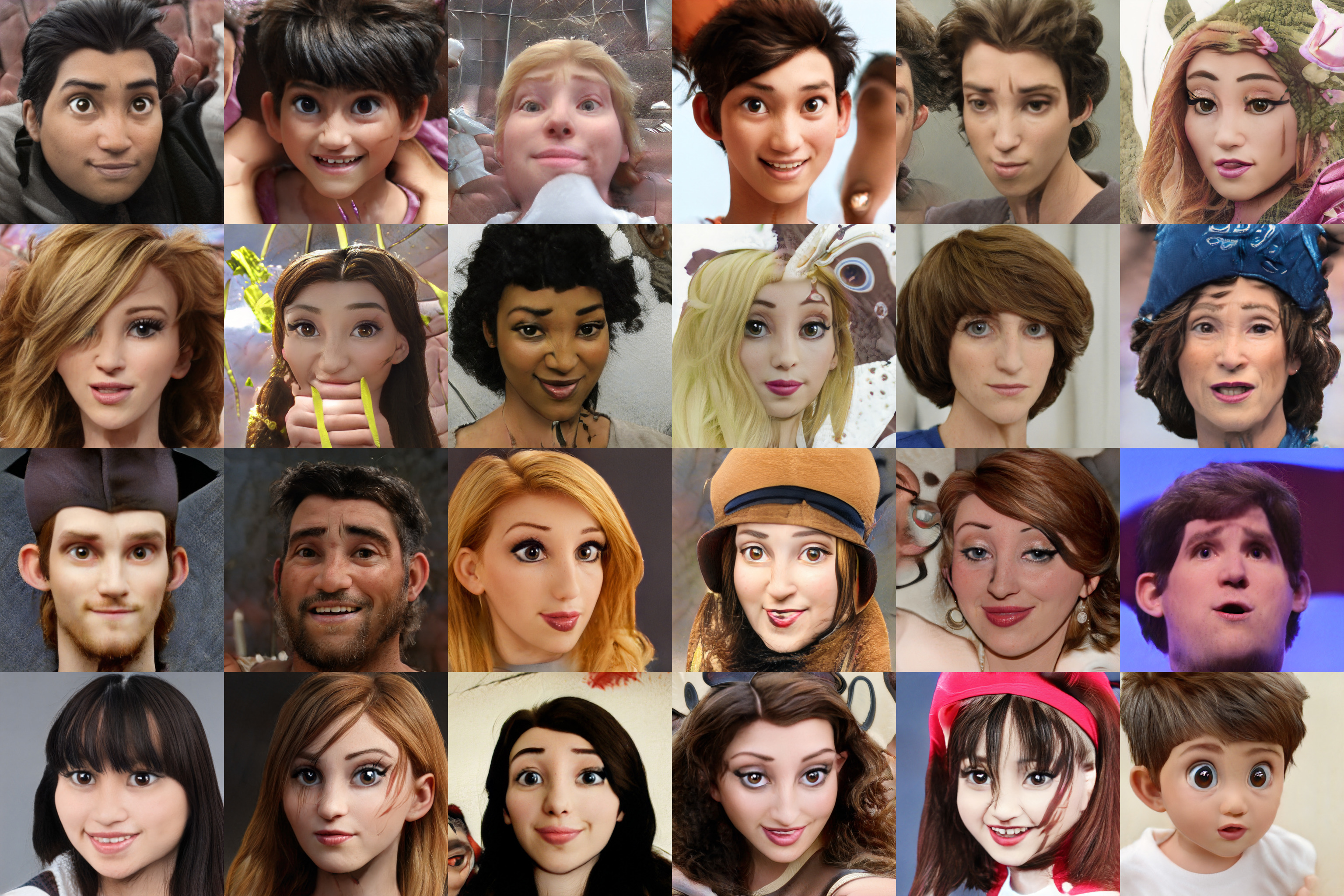}
  \caption{24 uncurated examples from the blended model shown in Figure \ref{fig:toonify}}
\end{figure}

\section{Further work}

In this work we demonstrated the simplest resolution dependent interpolation scheme of "layer swapping" but the interpolation function is an arbitrary choice. A further avenue for exploration would be interpolation schemes which vary smoothly according to resolution or those which target a certain resolution specifically, or those which involve more than two models. 

A logical extension of \emph{resolution} dependent interpolation is \emph{channel} dependent interpolation, where the interpolation is a function of both resolution and channel index in a particular convolutional layer. This could be used to target specific regions or features in the image, as it has been shown that certain channels affect particular regions of the generated image \cite{collins2020editing}. This would be a similar idea to the conditional interpolation introduced by \citet{collins2020editing} whereas here we would be interpolating between \emph{models} rather than latent vectors.

\end{document}